\documentclass{article}

\usepackage{arxiv}

\usepackage[T1]{fontenc}    
\usepackage{hyperref}       
\usepackage{url}            
\usepackage{booktabs}       
\usepackage{amsfonts}       
\usepackage{amsmath}
\usepackage{nicefrac}       
\usepackage{microtype}      
\usepackage{lipsum}		
\usepackage{graphicx}
\usepackage{natbib}
\usepackage{doi}
\usepackage[utf8]{inputenc}
\usepackage[english]{babel}

\usepackage[frozencache,cachedir=.]{minted}
\usepackage{listings}
\usepackage{multirow}
\usepackage{bm}

\usepackage{sidecap}   
\sidecaptionvpos{figure}{t} 

\def\vx{{\bm{x}}}

\def\mA{{\bm{A}}}

\def\mD{{\bm{D}}}

\def\mI{{\bm{I}}}

\def\mW{{\bm{W}}}
\def\mX{{\bm{X}}}

\def\mZ{{\bm{Z}}}

\def\gL{{\mathcal{L}}}

\def\gT{{\mathcal{T}}}

\def\sR{{\mathbb{R}}}

\title{Jointly Learnable Data Augmentations for Self-Supervised GNNs}

\author{{Zekarias T. Kefato} \\
	KTH Royal Institute of Technology\\
	Stockholm, Sweden \\
	\texttt{zekarias@kth.se} \\
	\And
	{Sarunas Girdzijauskas} \\
	KTH Royal Institute of Technology\\
	Stockholm, Sweden \\
	\texttt{sarunasg@kth.se} \\
	\And
	{Hannes St{\"a}rk} \\
	Technical University of Munich\\
	Munich, Germany \\
	\texttt{hannes.staerk@tum.de} \\
}

\newcommand{\model}{\textsc{GraphSurgeon}}

\newcommand{\gssl}{\textsc{ssl-gnn}}

\newcommand{\gcn}{\textsc{GCN}}
\newcommand{\gat}{\textsc{GAT}}
\newcommand{\sage}{\textsc{GraphSage}}

\newcommand{\cluster}{\textsc{ClusterGCN}}
\newcommand{\saint}{\textsc{GraphSaint}}
\newcommand{\pprgo}{\textsc{PPRGO}}

\newcommand{\dgi}{\textsc{DGI}}
\newcommand{\mvgrl}{\textsc{MVGRL}}
\newcommand{\gca}{\textsc{GCA}}
\newcommand{\selfgnn}{\textsc{SelfGNN}}
\newcommand{\bgrl}{\textsc{BGRL}}

\newcommand{\byol}{\textsc{BYOL}}

\begin{document}
\maketitle
\begin{abstract}

Self-supervised Learning (SSL) aims at learning representations of objects without relying on manual labeling.
Recently, a number of SSL methods for graph representation learning have achieved performance comparable to SOTA semi-supervised GNNs.
A Siamese network, which relies on data augmentation, is the popular architecture used in these methods.
However, these methods rely on heuristically crafted data augmentation techniques.
Furthermore, they use either contrastive terms or other tricks (e.g., asymmetry) to avoid trivial solutions that can occur in Siamese networks.

In this study, we propose,~\model, a novel SSL method for GNNs with the following features.
First, instead of heuristics we propose a learnable data augmentation method that is jointly learned with the embeddings by leveraging the inherent signal encoded in the graph.
In addition, we take advantage of the flexibility of the learnable data augmentation and introduce a new strategy that augments in the embedding space, called post augmentation.
This strategy has a significantly lower memory overhead and run-time cost.
Second, as it is difficult to sample truly contrastive terms, we avoid explicit negative sampling.
Third, instead of relying on engineering tricks, we use a scalable constrained optimization objective motivated by Laplacian Eigenmaps to avoid trivial solutions.

To validate the practical use of \model, we perform empirical evaluation using 14 public datasets across a number of domains and ranging from small to large scale graphs with hundreds of millions of edges.
Our finding shows that~\model~is comparable to six SOTA semi-supervised and on par with five SOTA self-supervised baselines in node classification tasks. The source code is available at \href{https://github.com/zekarias-tilahun/graph-surgeon}{https://github.com/zekarias-tilahun/graph-surgeon}.

\end{abstract}



\keywords{GNN, Self-Supervised Learning, Learnable Data Augmentation}



\section{Introduction}
\label{sec:intro}

Owing to its comparable performance to semi-supervised learning, self-supervised learning (SSL) has been widely adapted across a number of domains. 
In SSL, we seek to learn representations of objects (e.g. images and graphs) without relying on manual labeling.
A particular interest in this study is SSL for graph neural networks (GNN) -~\gssl.

Earlier efforts in SSL concentrated on devising a pretext task and training a model to extract transferable features.
However, it is challenging to learn a generalized representation that is invariant to the pretext task and useful for a downstream task~\cite{DBLP:journals/corr/abs-1912-01991}.
As a result, a new framework based on a Siamese network~\cite{10.5555/2987189.2987282} has become the de facto standard, and current SOTA results are achieved using different flavors of such networks~\cite{DBLP:journals/corr/abs-2006-09882,DBLP:journals/corr/abs-1912-01991,DBLP:journals/corr/abs-1911-05722,DBLP:journals/corr/abs-2002-05709,DBLP:journals/corr/abs-2006-07733,DBLP:journals/corr/abs-2104-14294,DBLP:journals/corr/abs-2103-03230,DBLP:journals/corr/abs-2103-14958,10.1145/3442381.3449802,DBLP:journals/corr/abs-2010-13902,DBLP:journals/corr/abs-2006-05582,DBLP:journals/corr/abs-2106-07594}.

Self-supervised techniques using Siamese networks learn representations that benefit from data-augmentation (perturbation)~\cite{DBLP:journals/corr/abs-2103-03230}, and devising suitable augmentation techniques is one of the main challenges.
Although there are well established data-augmentation techniques (e.g., for images), there are no standard techniques for other modalities, such as text and graph data.
As a result, existing methods depend on different heuristics and/or trial and error~\cite{DBLP:journals/corr/abs-2106-07594}.

In a Siamese framework, there are two networks, commonly referred to as a \textit{student} and \textit{teacher} network, and their inputs are differently augmented views of the same object.
Learning is carried out by maximizing the agreement between the outputs of the two networks.
However, a contrastive term (negative samples) is required to prevent a trivial solution (e.g., constant output, collapse)~\cite{DBLP:journals/corr/abs-2002-05709,zhu2020deep,DBLP:journals/corr/abs-1911-05722,sun2021mocl,10.1145/3442381.3449802,DBLP:journals/corr/abs-2006-05582,DBLP:journals/corr/abs-2106-07594,DBLP:journals/corr/abs-2006-09882}.
On the other hand, because it is usually difficult to sample truly contrastive (negative) terms, techniques were proposed without the need for explicit negative samples.
These methods rely on engineering tricks (e.g., asymmetry) to avoid a trivial solution~\cite{DBLP:journals/corr/abs-2006-07733,DBLP:journals/corr/abs-2011-10566,DBLP:journals/corr/abs-2104-14294,DBLP:journals/corr/abs-2103-14958,thakoor2021bootstrapped}.
Three strategies are commonly used, which are asymmetry in the (1) architecture (2) weights, and (3) update rules~\cite{DBLP:journals/corr/abs-2006-07733}.


In this study, we propose an SSL model for GNNs based on the Siamese architecture called~\model~ (\textbf{s}elf-s\textbf{u}pe\textbf{r}vised \textbf{G}NN that jointly l\textbf{e}arns t\textbf{o} augme\textbf{n}t).
Unlike most existing methods,~\model~ requires neither data augmentation using heuristics nor explicit negative samples.
We design an SSL architecture in such a way that data augmentation is jointly learned with the graph representation.
Moreover, by using a principled constrained optimization objective, we avoid the need for explicit negative samples.

To augment a given node $v$, we simply use two augmentation heads $v_1 = f_{\theta_1}(v)$ and ${v_2 = f_{\theta_2}(v)}$, parameterized by two separate sets of weights $\theta_1$ and $\theta_2$, which produce two views $v_1$ and $v_2$ of $v$. 
For example, $f_{\theta_1}(v)$ and $f_{\theta_2}(v)$ could be simple MLP heads.
$v_1$ and $v_2$ are fed to a shared GNN, $h_\theta(\cdot) $, parameterized by a set of weights $\theta$.
Here, the key idea is that the parameters of the augmentation heads, $\theta_1$ and $\theta_2$, are jointly learned with $\theta$.
Because the two networks are symmetric and equivalent, we use the terms \emph{left} and \emph{right}, instead of student and teacher.

\begin{SCfigure}
  \centering
  \includegraphics[width=0.5\textwidth]{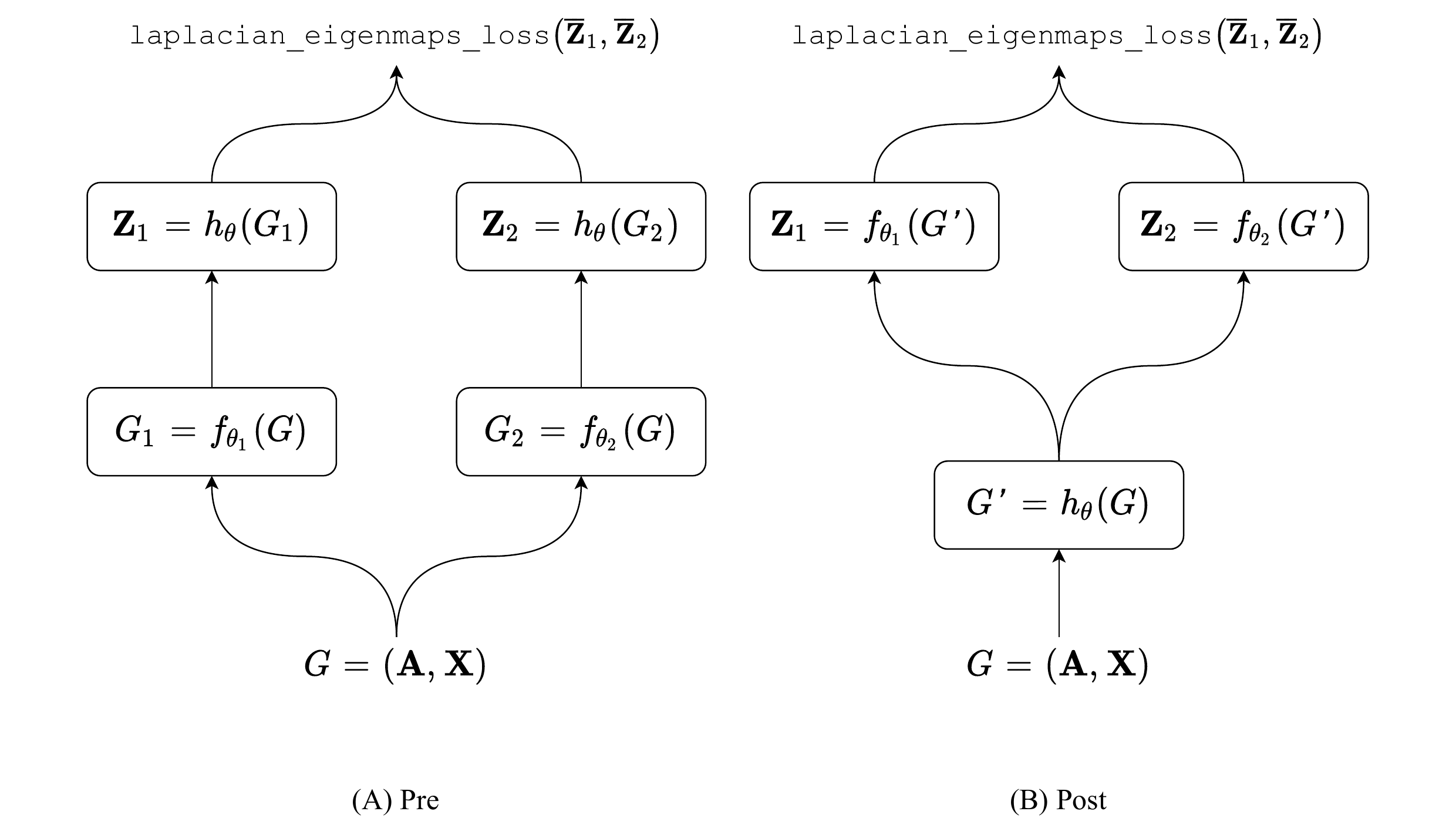}
  \caption{The architecture of~\model, (A) first generates two augmented views of an input signal from two learned functions $f_{\theta_1}$ and $f_{\theta_2}$ and then passes them to a shared GNN encoder $h_\theta$. Meanwhile, (B) first encodes the input signal and passes it to the two learned augmenters, $f_{\theta_1}$ and $f_{\theta_2}$.}
  \label{fig:graph_surgeon}
\end{SCfigure}

Furthermore, because of the flexibility of the learnable augmenters, we introduce an alternative new strategy called \emph{post-augmentation}.
Post-augmentation applies augmentation in a latent space.
That is, we first encode (e.g., using a GNN, CNN, or Transformer) and augment the encoded representations.
This is in contrast to the standard practice which we refer to as \emph{pre-augmentation}, where we first augment the input features and then encode them.
Compared to the pre-augmentation strategy, we show that post-augmentation significantly decreases memory use and computation time.
Fig.~\ref{fig:graph_surgeon} shows the pre and post architectures.
Note that post-augmentation is difficult, if not impossible, when using predefined augmentation strategies.

\model~ is trained based on a loss function that draws inspiration from Laplacian Eigenmaps~\cite{6789755}. 
Just like any Siamese network, it has a term to maximize the agreement between the outputs of the left and right networks.
To avoid trivial solutions, instead of engineering tricks, we use an orthonormality constraint similar to Laplacian Eigenmaps.
The constraint encourages positive pairs to be similar while preventing trivial solutions.

Finally, we perform an empirical evaluation using publicly available datasets and compare~\model~with strong SOTA baselines on three types of node classification tasks.
The results show that~\model~is comparable to semi-supervised methods, on average just 2 percentage points away from the best performing ones.
However, it is on par with the~\gssl~baselines.

Our contributions are summarised as follows:
\begin{itemize}
    \item We propose a~\gssl~ called~\model~that jointly learns to embed and augment graph data.
    \item We introduce an alternative new augmentation strategy that happens in the latent space as opposed to the input space and it leads to a significant improvement in resource usage, regarding GPU memory and run-time.
    \item \model~learns meaningful representations based on a constrained optimization objective that has theoretical motivation as opposed to engineering tricks to prevent trivial solutions.
    \item We carried out extensive experiments using 14 publicly available datasets ranging from small to large, spanning a number of domains and we show that~\model~is comparable to six SOTA supervised GNNs and on par with five SOTA~\gssl~architectures.
\end{itemize}

To the best of our knowledge, this is the first attempt towards~\gssl~that jointly learns data augmentation and also does not require explicit contrastive terms.
In addition, one can easily adopt the techniques for other domains, such as CV.

\section{Related Work}
\label{sec:related_work}

We briefly discuss~\gssl~ methods based on Siamese networks in terms of their data-augmentation strategies and their architecture choices.
Then we give an overview of GNNs. 

\paragraph{Data Augmentation} Although there are well-established data augmentation techniques in the CV domain, this is not the case for the graph domain~\cite{DBLP:journals/corr/abs-2006-05582,DBLP:journals/corr/abs-2106-07594}.
Different heuristics, based on high-order networks, perturbation of topology and attributes have been proposed~\cite{velickovic2018deep,DBLP:journals/corr/abs-2006-05582,DBLP:journals/corr/abs-2010-13902,DBLP:journals/corr/abs-2103-14958,bielak2021graph}.
It is not clear what the relative benefit of these augmentation strategies is, and little is known regarding the relevance of each strategy with respect to different downstream tasks.
There are also conflicting observations regarding the composition of more than two augmentation strategies~\cite{DBLP:journals/corr/abs-2006-05582,DBLP:journals/corr/abs-2106-07594}.
A recent study~\cite{DBLP:journals/corr/abs-2106-07594} proposes a different strategy that dynamically chooses augmentation techniques while learning the graph representation through a joint optimization objective.

Our study differs from this line of research, first as there are no predefined data augmentations whatsoever, and second, the augmentation is jointly learned (\emph{not chosen}) with the graph representation.

\paragraph{Architectures}
The key difference between existing architectures arises from the need to prevent trivial solutions.
To this end, existing studies~\cite{velickovic2018deep,DBLP:journals/corr/abs-2006-05582,DBLP:journals/corr/abs-2010-13902,10.1145/3442381.3449802,DBLP:journals/corr/abs-2106-07594,sun2021mocl,wang2021selfsupervised,zhu2020deep} often rely on contrastive architectures.
However, as sampling truly contrastive terms is a difficult task, other studies have used asymmetric architectures to prevent trivial solutions.
Originally proposed for CV~\cite{DBLP:journals/corr/abs-2006-07733,DBLP:journals/corr/abs-2011-10566}, asymmetric methods~\cite{thakoor2021bootstrapped,DBLP:journals/corr/abs-2103-14958,DBLP:journals/corr/abs-2104-14294} have empirically shown that having different student and teacher networks and a stop gradient operation is sufficient to prevent collapse.
Though there are three possibilities for asymmetry, not all of them are necessary to prevent trivial solutions~\cite{DBLP:journals/corr/abs-2011-10566,DBLP:journals/corr/abs-2104-14294}

In a similar line of research, a recent study~\cite{DBLP:journals/corr/abs-2103-03230,bielak2021graph} has introduced the notion of redundancy reduction to prevent collapse without requiring the asymmetry trick.
Our study advocates a similar notion as~\cite{DBLP:journals/corr/abs-2103-03230,bardes2021vicreg}, in the sense that it has no explicit contrastive terms and does not use the asymmetry trick.
On the other hand, we maximize agreement between positive pairs (as opposed to redundancy reduction), and to avoid collapse we capitalize on a constrained optimization objective motivated by Laplacian Eigenmaps~\cite{6789755}.


\paragraph{Graph Neural Network (GNN)}
Because there is a plethora of GNN architectures, our discussion focuses on a brief description of the essentials of GNNs.
Unlike previous neural network architectures, such as RNNs and CNNs, GNNs are highly flexible in terms of the type of data structure they can accommodate.
The GNN formulation allows us to learn representations of objects for a multitude of data structure, such as grids, graphs, groups, geodesics and gauges~\cite{bronstein2021geometric,9205745}.
Contrary to the rigid computational graphs of RNN and CNN, in a GNN it is the input data that determines the computation graph.
A GNN can be viewed as a message passing framework~\cite{gilmer2017neural}, where the messages  are vectors that are updated by a neural network head~\cite{9205745}, which is commonly shared by all nodes.
The propagation rule for the ``vanilla" GNN (GCN)~\cite{kipf2017semisupervised} is defined as

\begin{equation}\label{eq:gcn}
    \mX^{(l + 1)} = \sigma(\tilde{\mA}\mX^{(l)}\mW^{(l)})
\end{equation}

where $\sigma$ is an activation function, e.g., \texttt{ReLU}, $\tilde{\mA}$ is a symmetrically normalized adjacency matrix, $\mX$ is node feature matrix ($\mX^{(0)} = \mX$), $\mX^{(l)}$ is the node feature at the $l^{th}$ layer, and $\mW^{(l)}$ is the weight of the $l^{th}$ layer.

Because of the limitations of the above GNN model, such as scalability and oversmoothing, several extensions have been proposed~\cite{hamilton2018inductive,Chiang_2019,zeng2020graphsaint,wu2019simplifying,chen2018fastgcn,xu2018representation,chen2019measuring,zhou2020deeper,oono2021graph,Bojchevski_2020,gilmer2017neural}.
Of particular interest to this study are architectures for large scale graphs.
The popular ones are neighborhood sampling~\cite{hamilton2018inductive,chen2018fastgcn} and subgraph sampling ~\cite{Chiang_2019,zeng2020graphsaint}.
Instead of a full-batch training, which is used in the ``vanilla" architecture, a neighborhood (layer) sampling  uses a sampled set of neighbors for message passing.
On the other hand, subgraph sampling techniques in a nutshell build sub-graphs as batches and apply a full-batch GNN over subgraphs (using clustering~\cite{Chiang_2019} or subgraph sampling~\cite{zeng2020graphsaint}).



\section{\model}
\label{sec:surgeon}
We consider an undirected graph $G$ with a set of  nodes  $V$ and edges $E$, where $N = \vert V \vert$ and $M = \vert E \vert$.
The adjacency matrix representation of $G$ is given by $\mA \in [0, 1]^{N \times N}$; and ${\tilde{\mA} = \mD^{1/2} (\mA + \mI_N) \mD^{1/2}}$ denotes the symmetrically normalized adjacency matrix. 
Where ${\mD_{i,i} = \sum_{j} \mA_{i, j} + 1}$ is the degree matrix and $\mI_N$ is and Identity matrix of $N$ diagonal entries.
Each node $u$ is associated with an attribute signal $\vx_u \in \sR^F$, and the set of signals for all the nodes in the graph is given by $\mX \in \sR^{N \times F}$.

Though the proposed method is agnostic to the type of GNN architecture, we assume a kind of GNN, parameterized by $\theta = \lbrace \mW^{(l)} \in \sR^{F_{l - 1} \times F_{l}}: l=1, \ldots , L \rbrace$ is given. For example, the GCN model by Kipf and Welling~\cite{kipf2017semisupervised} with the propagation rule in Eq.~\ref{eq:gcn}.
Hence, we consider a generic $L$--layer message passing GNN, $h_\theta$, parameterized by $\theta$.
Similar to Eq.~\ref{eq:gcn}, at every layer $l$ in $h_\theta$, each node propagates messages to its neighbors using a particular materialization of $G$, e.g., $\tilde{\mA}$.
To add expressivity, the propagation is followed by a linear transformation using $\mW^{(l)}$ and then a non-linearity using \texttt{ReLU}.

Given a GNN $h_\theta$, a certain materialization of $G$, and a node feature matrix $\mX$, we propose,~\model, a novel~\gssl~model based on the Siamese network~\cite{10.5555/2987189.2987282}.
The overview of the model is shown in Fig.~\ref{fig:graph_surgeon}.
For ease of discussion, we consider $\mA$ as the materialization of G.
We express the GNN encoder using two equivalent formulations, $h_\theta(\mA, \mX) $ and $h_\theta(G)$, interchangeably.


\subsection{Learning Data Augmentation}
\label{sub:sec:learning_data_augmentation}
The core component behind SSL models based on a Siamese network is data augmentation, without it, such models have poor empirical performance~\cite{zhu2020deep}.
However, existing augmentation strategies are often based on heuristics and/or trial and error~\cite{DBLP:journals/corr/abs-2106-07594}.
That is, given a graph $G$, two augmented views $G_1 = t_1(G)$ and $G_2 = t_2(G)$ are generated by applying augmentation techniques $t_1 \sim \gT$ and $t_2 \sim \gT$, sampled from a set of predefined techniques, $\gT$.

Automatically learning augmentations has not been well explored yet.
In this study, we propose a simple yet novel data augmentation technique that is learned based on the signal encoded in the graph.
Therefore, we replace $t_1$ and $t_2$, with trainable functions $f_{\theta_1}$ and $f_{\theta_2}$ parametrized by $\theta_1$ and $\theta_2$.

One can model $f$ to learn either a different view of the topology $\mA$, or attribute signal $\mX$.
In this study, we address the latter, and the former will be covered in future work.
Therefore, we model $f$ as an MLP, \emph{i.e.}, two augmented views are generated as $\mX_1 = f_{\theta_1}(\mX)$, $\mX_2 = f_{\theta_2}(\mX)$, where $\mX_1 \in \sR^{N \times D}$ and $\mX_2 \in \sR^{N \times D}$ and $\theta_1 = \lbrace \mW_1^{(l)}: l=1, \ldots L_1 \rbrace$ and $\theta_2 = \lbrace \mW_2^{(l)}: l = 1, \ldots L_2 \rbrace$, where $L_1$ and $L_2$ are the number of layers of $f_{\theta_1}$ and $f_{\theta_2}$.
In order for the graph signal to govern the learned augmentations $\mX_1$ and $\mX_2$, the key design choice in~\model~is to jointly learn $\theta_1$ and $\theta_2$ with $\theta$.

A standard~\gssl~technique follows a \emph{pre-augmentation} architecture (in short, \emph{pre}), i.e. first the input is augmented and then fed to a GNN encoder, Fig~\ref{fig:graph_surgeon} (A).
Our design, however, gives us the flexibility to change the order of augmentation to \emph{post-augmentation} (in short, \emph{post}), such that we first encode and then augment, Fig.~\ref{fig:graph_surgeon} (B).
The \emph{post} architecture is motivated by efficiency in terms of run time and memory footprint.

While both architectures have the same number of parameters, \emph{post} requires less computation time and memory during training. This is the case since with \emph{pre}, the GNN encoder processes two separate representations for each node, leading to twice as many computations for the encoder. In \emph{post}, the encoder only processes one representation for each node and the augmentation happens afterwards. With the GNN encoder making up the largest part of the architecture, this means that \emph{pre} needs almost two times more resources than \emph{post}.



\subsection{Joint Training of the Parameters}
\label{sub:sec:joint_training}

\begin{listing}[ht!]
\begin{minted}
[
frame=lines,
framesep=2mm,
baselinestretch=1.2,
fontsize=\footnotesize,
]
{python3}

# h: GNN Encoder
# f_1: First augmenter
# f_2: Second augmenter
# gamma: weight for the regularization
# B: Number of nodes in a batch
# D: Number of features after augmentation
# F_l: The final size of the embedding dimension

for data in loader:
    # data contains the graph data, that is, adj: a sparse adjacency 
    # matrix and x: the node features. It could be a full-batch data
    # or a block-diagonal of b batches of subgraphs with the 
    # corresponding node features.

    if pre_augment:
        # data.x.shape = BxF
        # data.adj.shape = BxB
        x1 = f_1(data.x) # x1.shape = BxD
        x2 = f_2(data.x) # x2.shape = BxD
        z1 = h(data.adj, x1) # z1.shape = BxF_l
        z2 = h(data.adj, x2) # z2.shape = BxF_l
    else:
        z = h(data.adj, data.x) # z.shape = BxD
        z1 = f_1(z) # z1.shape = BxF_l
        z2 = f_2(z) # z2.shape = BxF_l
        
    # Transforming embedding vectors to unit vectors
    z1_unit = torch.nn.functional.normalize(z1, dim=1, p=2) 
    z2_unit = torch.nn.functional.normalize(z2, dim=1, p=2) 
    
    mse = mse_loss(z1_unit, z2_unit) # Equation 2
    if use_improved_loss:
        I = torch.eye(F_l)
        constraint = (
            (z1_unit.t().matmul(z1_unit) - I).norm(p='fro') +
            (z2_unit.t().matmul(z2_unit) - I).norm(p='fro')
        ) # Column orthonormality constraint
    else:
        I = torch.eye(B)
        constraint = (
            (z1_unit.matmul(z1_unit.t()) - I).norm(p='fro') +
            (z2_unit.matmul(z2_unit.t()) - I).norm(p='fro')
        ) # Row orthonormality constraint
        
    optimizer.zero_grad()
    loss = mse + gamma * constraint # Equation 4
    loss.backward()
    optimizer.step()
    
\end{minted}
\caption{A PyTorch and PyTorch Geometric based pseudo code for training \model}
\label{lst:surgeon_pseudo_code}
\end{listing}

The following discussion assumes the pre-augmentation architecture.
The high-level overview for the training function of~\model~is given in Listing~\ref{lst:surgeon_pseudo_code}.
\model~ first generates augmented views $G_1 = (\mA, \mX_1) = f_{\theta_1}(G)$ and $G_2 = (\mA, \mX_2) = f_{\theta_2}(G)$, $\texttt{f}\_1\texttt{(data.x)}$ and $\texttt{f}\_2\texttt{(data.x)}$ in the pseudocode. 
Then it encodes the views using a shared GNN, $h_\theta$, to generate the corresponding latent representations $\mZ_1=h_\theta(G_1)$ and $\mZ_2=h_\theta(G_2)$ ($\texttt{h}\texttt{(data.adj, x1)}$ and $\texttt{h}\texttt{(data.adj, x2)}$ in the listing).

Our goal in an \gssl~framework is to maximize the agreement between these two representations.
To this end, we closely follow Laplacian Eigenmaps~\cite{6789755} and minimize the mean squared error between the normalized representations (unit vectors) of two data points.
Though in Laplacian Eigenmaps the two data points correspond to different objects (e.g., two different nodes, images), in our case, these are just the unit embedding vectors of the two augmented views, which are $\bar{\mZ}_1$ and $\bar{\mZ}_2$ ($\texttt{z1}\_\texttt{unit}$ and $\texttt{z2}\_\texttt{unit}$ in the pseudocode).
Therefore, we define the objective based on these unit vectors as:
\begin{equation}\label{eq:unconstrained_objective}
    \gL_\theta = \vert \vert \bar{\mZ}_1 - \bar{\mZ}_2 \vert \vert^2_F
\end{equation}
However, Eq.~\ref{eq:unconstrained_objective} admits a trivial solution, that is, collapse into a single point or a subspace~\cite{6789755}.
For this reason, we modify E.q.~\ref{eq:unconstrained_objective} and incorporate an orthonormality constraint inspired by the Laplacian Eigenmaps.
Moreover, we want to jointly optimize the parameters of the augmenters. 
To achieve these goals we update E.q.~\ref{eq:unconstrained_objective} and formulate it as a constrained optimization objective as in Eq.~\ref{eq:constrained_objective}.
\begin{gather}\label{eq:constrained_objective}
    \gL_{\theta, \theta_1, \theta_2} = \vert \vert \bar{\mZ}_1 - \bar{\mZ_2} \vert \vert^2_F \\
    s.t. \quad \bar{\mZ_1}\bar{\mZ_1}^T = \mI_N \quad and \quad \bar{\mZ_2}\bar{\mZ_2}^T = \mI_N \nonumber
\end{gather}
Eq.~\ref{eq:constrained_objective} encourages positive pairs across $\bar{\mZ_1}$ and $\bar{\mZ_2}$ to be similar to each other, and the orthonormality constraint ensures that each row in $\bar{\mZ_1}$ or $\bar{\mZ_2}$ is similar to itself and orthonormal to other rows.
Consequently, a trivial solution is avoided~\cite{6789755}.
As stated earlier, existing methods often rely on contrastive terms or the asymmetry trick to achieve this.

By using the Lagrangian, we can relax the constrained optimization problem using Eq.~\ref{eq:final_objective}, to obtain a regularized objective which we call Laplacian Eigenmaps loss 
\begin{equation}\label{eq:final_objective}
    \gL_{\theta, \theta_1, \theta_2} = \vert \vert \bar{\mZ}_1 - \bar{\mZ_2} \vert \vert^2_F + \gamma  \big(\vert \vert \bar{\mZ_1}\bar{\mZ_1}^T - \mI_N\vert \vert_F + \vert \vert \bar{\mZ_2}\bar{\mZ_2}^T - \mI_N \vert \vert_F \big)
\end{equation}


\subsubsection*{Improved objective}
Although E.q.~\ref{eq:final_objective} addresses the aforementioned concerns, the matrix multiplications, $\bar{\mZ_1}\bar{\mZ_1}^T$ and $\bar{\mZ_2}\bar{\mZ_2}^T$, produce $N \times N$ dense matrices.
This is not desirable for full-batch GNNs, since storing the resulting matrix in a GPU memory is not feasible.
For this reason, we improve the regularization by making a simple change and replacing the row orthonormality regularization with column \emph{orthogonality} regularization. That is, replacing $\bar{\mZ_1}\bar{\mZ_1}^T - \mI_N$ with $\bar{\mZ_1}^T\bar{\mZ_1} - \mI_{F_L}$ and $\bar{\mZ_2}\bar{\mZ_2}^T - \mI_N$ with $\bar{\mZ_1}^T\bar{\mZ_1} - \mI_{F_L}$.
Since $F_L \ll N$, is usually in the orders of hundreds, we can easily store an $F_L \times F_L$ matrix in the GPU's memory. 

Finally, when training~\model, gradients are back propagated on both the left and right networks.
As a result, all the parameters are updated according to the loss incurred with respect to the signal from the graph.
This in turn, allows the parameters of both the encoder, $h_\theta$, and the augmenters, $f_{\theta_1}$ and $f_{\theta_2}$, to be governed by the graph signal.
The fact that we have jointly trainable augmenters gives us the flexibility to use Fig.~\ref{fig:graph_surgeon} (A) and (B) without compromising the qualitative performance.
This is difficult, if not impossible, when using predefined augmentation techniques.


\subsection{Scalability}
\label{sub:sec:scalability}
The flexibility of~\model's architecture allows us to seamlessly integrate it into virtually any kind of GNN architecture.
For large-scale graphs with hundreds of millions of edges, one can integrate~\model~with existing methods for large-scale graphs.
For example, with methods based on neighborhood (layer) sampling~\cite{hamilton2018inductive} and subgraph sampling~\cite{Chiang_2019,zeng2020graphsaint}.
For small graphs, we use full-batch GCN~\cite{kipf2017semisupervised}, while for large-scale graphs we use neighborhood sampling unless stated otherwise.

\section{Experiments}
\label{sec:experiments}
\begin{SCtable}[]
\centering
\begin{tabular}{l|l|l|l|l|l}
\hline
\textbf{Dataset} & $N$ & $M$ & $F$ & $C$ & Task \\ \hline
Cora Full & 19,793 & 126,842 & 8,710 & 70 & $MCC$\\ \hline
DBLP & 17,716 & 105,734 & 1,639 & 4 & $MCC$\\ \hline
PubMed & 19,717 & 88,648 & 500 & 3 & $MCC$\\ \hline
Physics & 34,493 & 495,924 & 8,415 & 5 & $MCC$\\ \hline
CS & 18,333 & 163,788 & 6,805 & 15 & $MCC$\\ \hline
Computers & 13,752 & 491,722 & 467 & 10 & $MCC$\\ \hline
Photo & 7,650 & 238,162 & 745 & 8 & $MCC$\\ \hline
Facebook & 22,470 & 342,004 & 128 & 4 & $MCC$\\ \hline
Flickr & 89,250 & 899,756 & 500 & 7 & $MCC$\\ \hline
GitHub & 37,700 & 578,006 & 128 & 2 & $BC$\\ \hline
WikiCS & 11,701 & 297,110 & 300 & 10 & $MCC$\\ \hline
Actor & 7,600 & 30,019 & 932 & 5 & $MCC$\\ \hline \hline
Yelp & 716,847 & 13,954,819 & 300 & 100 & $MLC$\\ \hline
Reddit & 232,965 & 114,615,892 & 602 & 41 & $MCC$\\ \hline
\end{tabular}
\caption{Summary of the datasets, and $N = \vert V \vert$, $M = \vert E \vert$, $F$ is the number of features, and $C$ is the number of classes. $BC$, $MCC$ and $MLC$ represent binary, multi-class and multi-label classification, respectively}
\label{tbl:datasets}
\end{SCtable}

We validate the practical use of~\model~using 14 publicly available datasets, ranging from small to large-scale graphs.
All of the datasets are collected from PyTorch Geometric (PyG)~\footnote{https://pytorch-geometric.readthedocs.io/en/latest/index.html}, and grouped as
\begin{itemize}
    \item Citation Networks (\emph{Cora}, \emph{DBLP}, and \emph{PubMed}): Paper to paper citation networks, and we classify papers into different subjects~\cite{hamilton2018inductive}.
    \item Co-Author Networks (Computer Science (\emph{CS}) an \emph{Physics}): Author collaboration network from Microsoft Academic Graph, and the task is to predict the active field of authors~\cite{shchur2019pitfalls}.
    \item Co-Purchased Products Network (\emph{Computers} and \emph{Photo}): Co-purchased products from the respective categories on Amazon, and the task is to predict the refined categories~\cite{shchur2019pitfalls}.
    \item Wikipedia (\emph{Actor} and \emph{WikiCS}): WikiCS contains Wikipedia hyperlinks between Computer Science articles, and we classify articles into branches of CS~\cite{shchur2019pitfalls}, and Actor contains actors co-occurrence on the same Wikipedia article and we classify actors into groups based word of actors' Wikipedia~\cite{pei2020geomgcn}.
    \item Social (\emph{Facebook}, \emph{Flickr}, \emph{GitHub}, \emph{Reddit}, and \emph{Yelp}):
    Facebook contains a page to page graph of verified Facebook sites, and we want to classify pages into their categories~\cite{rozemberczki2021multiscale}. Flickr contains a network of images based on common properties (e.g., geo-location) along with their description, and the task is to predict a unique tag of an image~\cite{zeng2020graphsaint}. 
    GitHub contains the social network of developers, and we want to classify developers as web or machine learning developers~\cite{rozemberczki2021multiscale}. Yelp is also the social network of Yelp users, and we predict business categories each user has reviewed. For Reddit, we predict the subreddits (communities) of user posts~\cite{hamilton2018inductive,zeng2020graphsaint}. 
\end{itemize}
A brief summary of the datasets is provided in Table~\ref{tbl:datasets}.
We also group them into two, as large (Yelp and Reddit) and small (the rest).

We compare~\model~against 11 state-of-the-art baselines grouped into two
\begin{itemize}
    \item \emph{Semi-Supervised}: Six of the baselines are methods that use a fraction of the node labels during training, three of which (\gcn~\cite{kipf2017semisupervised}, \gat~\cite{velickovic2018graph}, \sage~\cite{hamilton2018inductive}) are used for small and medium-size graphs and the rest (\cluster~\cite{Chiang_2019}, \saint~\cite{zeng2020graphsaint}, and \pprgo~\cite{Bojchevski_2020}) for large-scale graphs.
    \item \emph{Self-supervised}: There are five methods under this group, three of which (\dgi~\cite{velickovic2018deep}, \mvgrl~\cite{DBLP:journals/corr/abs-2006-05582}, and \gca~\cite{10.1145/3442381.3449802}) use a contrastive architecture to prevent a trivial solution and the other two \selfgnn~\cite{DBLP:journals/corr/abs-2103-14958} and \bgrl~\cite{thakoor2021bootstrapped} use asymmetry. 
    Because these two techniques extend the same method, \byol~\cite{DBLP:journals/corr/abs-2006-07733}, for visual representation to graph representation and use the same code base, we present them as one.
\end{itemize}
For \gcn, \gat, \sage, \cluster, \saint, and \dgi~we use the implementation from PyG. 
For the rest we use the official implementation provided by the authors.

\subsection{Experimental Protocol}
For all the datasets we have three splits, training, validation and test.
For some of them, we use the splits provided by PyTorch Geometric, and for the rest, we randomly split them into 5\% training, 15\% validation and 80\% test sets.
We tune the hyperparameters of all the algorithms using Bayesian optimization~\footnote{We use OPTUNA for this purpose: https://optuna.readthedocs.io/en/stable/}, however for a fair comparison, we fix the representation dimension to 128.
In addition, we run all the models for 500 epochs and take the epoch with the best validation score.

We train the semi-supervised methods using the training split and tune their hyperparaters using the validation set.
Finally, we use the test set to infer the labels and report their performance.
For the self-supervised methods, we train them without any label and tune them using the validation set.
Following standard practice, they are evaluated under the linear protocol.
This means that we freeze the models and add a logistic regression (linear) classifier on top.
The linear classifier is trained using only the training split for 100 and 500 epochs, for small and large datasets, respectively.
Similar to the semi-supervised setting, we use the test set to simply predict the labels and report prediction quality.

We have three types of node classification tasks, which are binary, multi-class, and multi-label classifications.
Similar to existing studies, for the binary and multi-class tasks, we use accuracy, and for the multi-label, the Area Under the Receiver Operating Characteristic Curve (ROC-AUC).

Unless a different setting is stated, we assume the aforementioned protocol.

\subsection{Results}
\label{sec:results}
\begin{table*}[ht!]
\centering
\resizebox{\textwidth}{!}{
\begin{tabular}{|l|l|l|l||l|l|l||l||l|}
\hline
\multirow{3}{*}{\textbf{Datasets}} & \multicolumn{8}{c|}{\textbf{Algorithms}} \\ \cline{2-9} 
 & \multicolumn{3}{c||}{\textbf{Semi-Supervised}} & \multicolumn{3}{c||}{\textbf{Contrastive}} & \textbf{Asymmetric} & \multirow{2}{*}{\model} \\ \cline{2-8}
 & \gcn & \gat & \sage & \dgi & \mvgrl & \gca & \begin{tabular}[c]{@{}l@{}}\selfgnn/\\ \bgrl\end{tabular} &  \\ \hline \hline
Cora & \textbf{60.1$\pm$.001} & 58.27$\pm$.003 & 57.45$\pm$.003 & 50.66$\pm$.001 & 39.42$\pm$.193 & 37.64$\pm$.014 & 54.61$\pm$.135 & \textbf{56.33$\pm$.07} \\ \hline
DBLP & 82.7$\pm$.002 & \textbf{82.88$\pm$.002} & 81.39$\pm$.005 & 78.87$\pm$.002 & 69.2$\pm$.052 & 81.16$\pm$.007 & 81.32$\pm$.071 & \textbf{81.48$\pm$.09} \\ \hline
PubMed & \textbf{85.62$\pm$.001} & 84.98$\pm$.002 & 84.73$\pm$.001 & 84.28$\pm$.001 & 77.99$\pm$.315 & 82.76$\pm$.005 & 84.6$\pm$.076 & \textbf{84.94$\pm$.091} \\ \hline \hline
Physics & \textbf{95.4$\pm$.001} & 95.02$\pm$.002 & $\dagger$ & 94.92$\pm$.001 & 91.18$\pm$.024 & $\dagger$ & \textbf{95.11$\pm$.07} & \textbf{95.11$\pm$0.025} \\ \hline
CS & \textbf{91.87$\pm$.001} & 91.07$\pm$.002 & 91.44$\pm$.001 & 91.72$\pm$.001 & 87.18$\pm$.095 & 88.01$\pm$.005 & \textbf{92.23$\pm$.01} & 92.03$\pm$.0 \\ \hline \hline
Computers & \textbf{88.54$\pm$.003} & 88.3$\pm$.006 & 87.93$\pm$.004 & 80.28$\pm$.004 & 78.57$\pm$.14 & 74.04$\pm$.005 & \textbf{86.23$\pm$.139} & 85.16$\pm$.133 \\ \hline
Photo & 93.02$\pm$.003 & 93.18$\pm$002 & \textbf{93.64$\pm$.002} & 92.36$\pm$.06 & 86.04$\pm$.12 & 84.93$\pm$.009 & \textbf{92.87$\pm$.08} & 92.27$\pm$.05 \\ \hline \hline
Actor & 28.38$\pm$.008 & 28.62$\pm$.01 & \textbf{33.88$\pm$.007} & 29.93$\pm$.007 & \textbf{63.3$\pm$.03} & 27.39$\pm$.01 & 29.41$\pm$1.46 & 30.19$\pm$.34 \\ \hline
WikiCS & 76.87$\pm$.006 & 77.38$\pm$.005 & \textbf{77.41$\pm$.006} & 70.01$\pm$.007 & 61.7$\pm$.52 & 75.25$\pm$.006 & 75.34$\pm$.528 & \textbf{75.59$\pm$.11} \\ \hline \hline
Facebook & \textbf{89.5$\pm$.002} & 89.3$\pm$.01 & 89.25$\pm$.002 & 82.42$\pm$.001 & 78.88$\pm$0.045 & 86.29$\pm$0.004 & \textbf{86.38$\pm$.084} & 84.92$\pm$0.015 \\ \hline
Flickr & 51.66$\pm$.001 & 42.35$\pm$.001 & \textbf{52.11$\pm$.001} & 45.94$\pm$.001 & $\dagger$ & $\dagger$ & \textbf{51.26$\pm$.528} & 50.91$\pm$.054 \\ \hline
Github & \textbf{86.14$\pm$.001} & 86.16$\pm$.001 & 85.77$\pm$.001 & 83.84$\pm$.001 & 83.93$\pm$0.032 & $\dagger$ & 85.58$\pm$.053 & \textbf{85.7$\pm$.028} \\ \hline
\end{tabular}
}
\caption{The classification accuracy results along with the standard deviation. The bold highlight indicates the best performing algorithms from both the semi-supervised and self-supervised methods. $\dagger$ indicates that the algorithm has crashed because of an out-of-memory error.}
\label{tbl:small_medium_results}
\end{table*}

\begin{SCtable}[][ht!]
\centering
\begin{tabular}{|l|c|c|}
\hline
\multicolumn{1}{|c|}{\multirow{2}{*}{Algorithms}} & \multicolumn{2}{c|}{Datasets} \\ \cline{2-3} 
\multicolumn{1}{|c|}{}      & Yelp (ROC-AUC)  & Reddit (Accuracy) \\ \hline
        \cluster~(semi)           & 78.21 & 95.33  \\ \hline
        \saint~(semi)           & 75.62 & 95.73  \\ \hline
        \pprgo~(semi)           & 77.7  & 91.8   \\ \hline
        \model              & 77.44 & 91.22  \\ \hline
\end{tabular}
\caption{The prediction quality for the large scale datasets. For this experiment, we use semi-supervised and scalable GNN architectures as the full-batch ones do not fit in GPU memory. In addition, all the \gssl~baselines throw an out of memory error.}
\label{tbl:large_scale_results}
\end{SCtable}

The node classification results are reported in Tables~\ref{tbl:small_medium_results} and~\ref{tbl:large_scale_results}. 
Overall, \model~is comparable to the semi-supervised baselines and in-par (sometimes marginally better and at times marginally lower than) the self-supervised baselines.

As expected, the semi-supervised models are consistently better than the self-supervised ones, except for Actor.
\mvgrl~gives the best result, with more than $ 90\%$ improvement over the best performing method.
This comes as a result of using higher-order augmentation that happens to be beneficial for the Actor dataset.
A similar performance is not observed for~\mvgrl~on the other datasets.
This provides a motivation for automatically learning high-order topology signals that benefit some datasets.
As stated earlier, this will be covered in future work.

However, our finding showcases that just using the learned attribute augmentations and without requiring explicit negative samples, one can achieve a performance consistently close to semi-supervised models across a number of datasets and classification tasks.
On average, our model is at most 2 percentage points away from the best performing semi-supervised method.
Moreover, it scales to large networks with hundreds of millions of edges, where the other~\gssl~methods failed to handle.

\subsection{Ablation studies}
In the following, we investigate the impact of different aspects of \model.

\subsubsection{Loss Function}
\begin{SCfigure}[][t!]
    \centering
    \includegraphics[scale=0.37]{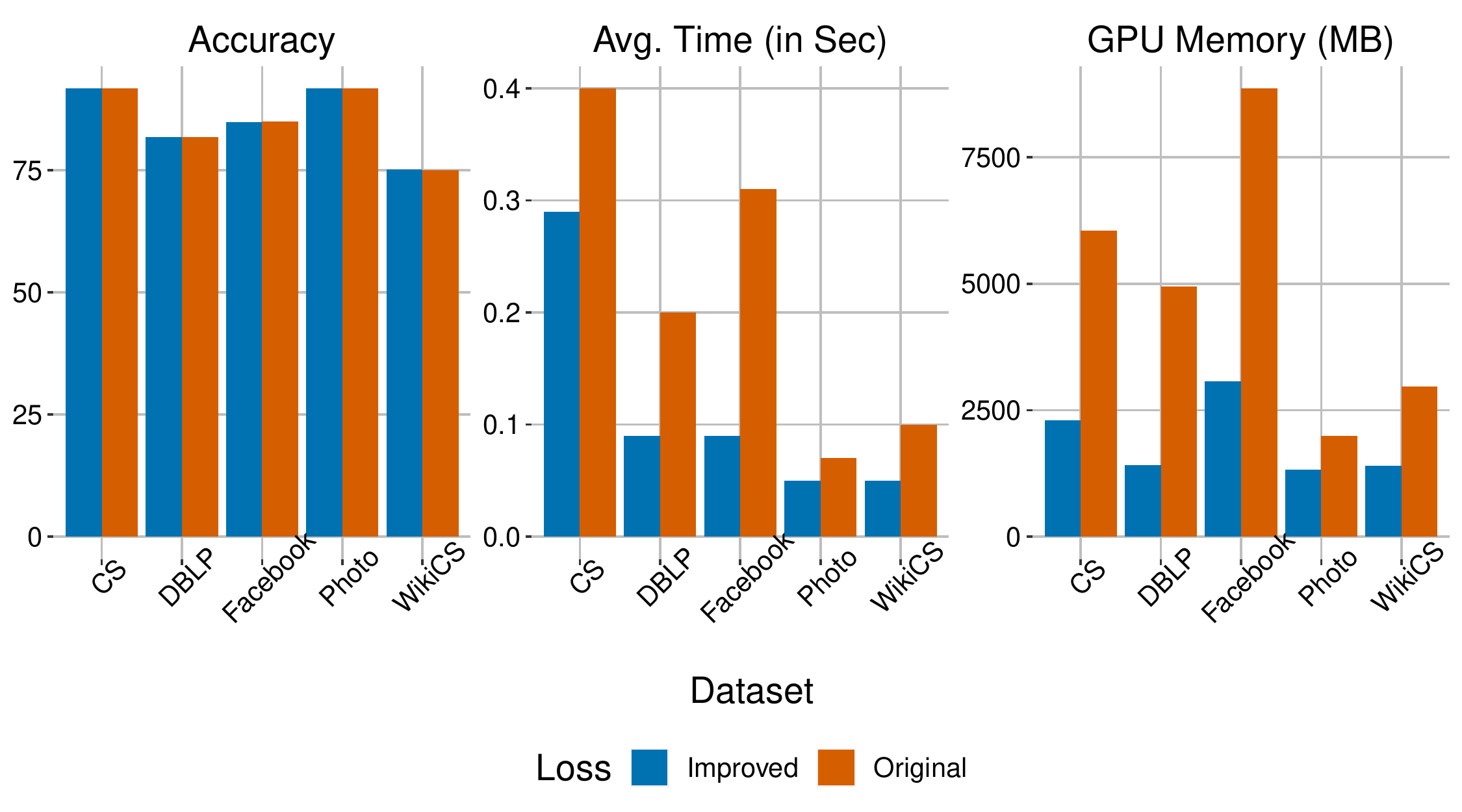}
    \caption{Comparison of the original and improved loss function in terms of accuracy, memory usage and run time (time to finish an epoch).}
    \label{fig:loss_ablation}
\end{SCfigure}

We have seen the constrained optimization objective in Section~\ref{sub:sec:joint_training} and also shown a way to improve it.
In the following, we analyze the effect of using the original vs. the improved loss function with respect to prediction accuracy and resource usage.
For the qualitative experiment, we train both flavors for 100 epochs and just 1 epoch otherwise.
The results of this experiment are reported in Fig~\ref{fig:loss_ablation}.
As expected, both flavors achieve similar qualitative performance.
Nonetheless, the improved version is significantly better than the original one in terms of memory usage and run time.

\subsubsection{Batch Size}
\begin{SCfigure}[][t!]
    \centering
    \includegraphics[scale=0.4]{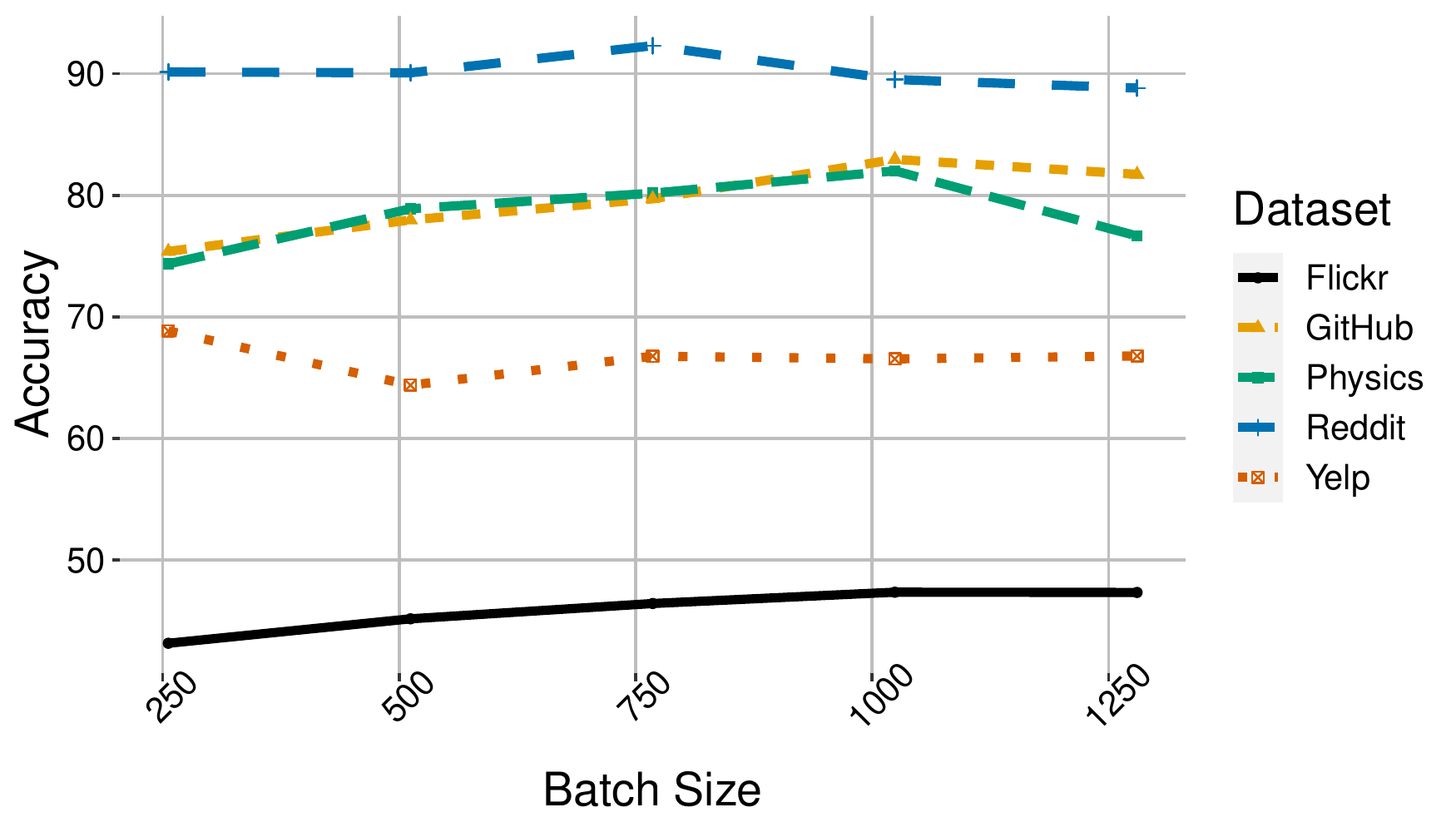}
    \caption{Effect of batch size on \model's performance}
    \label{fig:batch_size_effect}
\end{SCfigure}

As contrastive signals are indirectly injected because of the orthogonality constraint of Eq.~\ref{eq:final_objective}, it is important to analyze the impact of batch size to see if a large batch size is needed to effectively avoid trivial solutions.
For this reason, we train the model using sampled neighborhood subgraphs~\cite{hamilton2018inductive} instead of full-batch, and both the model and the linear head are trained for 100 epochs.
The results are reported in Fig.~\ref{fig:batch_size_effect}, and in general performance is directly proportional to batch size until a certain point.
For small datasets, there is improvement up to 1024.
The largest improvements are for the GitHub and Physics datasets, which are 7.57 (from 75.38\% to 82.95\%) and 7.66 perecentage points of accuracy (from 74.35\% to 82.01\%), respectively. 
However, that is not the case for the larger ones (Reddit and Yelp), where both smaller and larger batch sizes give comparable performance.
Overall, we have not observed qualitative differences for batch sizes bigger than 1024.
In our experiments, batch size greater than 1024 is only related to faster training, not improved quality.

\subsubsection{Embedding Size}
To provide a perspective, for this analysis we include the baselines. 
We use the same setting as the first experiment (Tables~\ref{tbl:small_medium_results} and~\ref{tbl:large_scale_results}), and we examine embedding sizes in $[256, 512, 768, 1024, 1280]$.
\model~and~\dgi~have the tendency to improve as we increase the embedding size.
On the other hand,~\gca~and~\selfgnn/\bgrl,~stay the same or decrease.
For \mvgrl, it seems that it has the tendency to improve proportional to the embedding size.
However, we were able to observe only for 256 and 512, as it throws an out of memory error for larger values.

\subsubsection{Pre vs. Post Augmentation}
\begin{figure}
    \centering
    \includegraphics[scale=0.45]{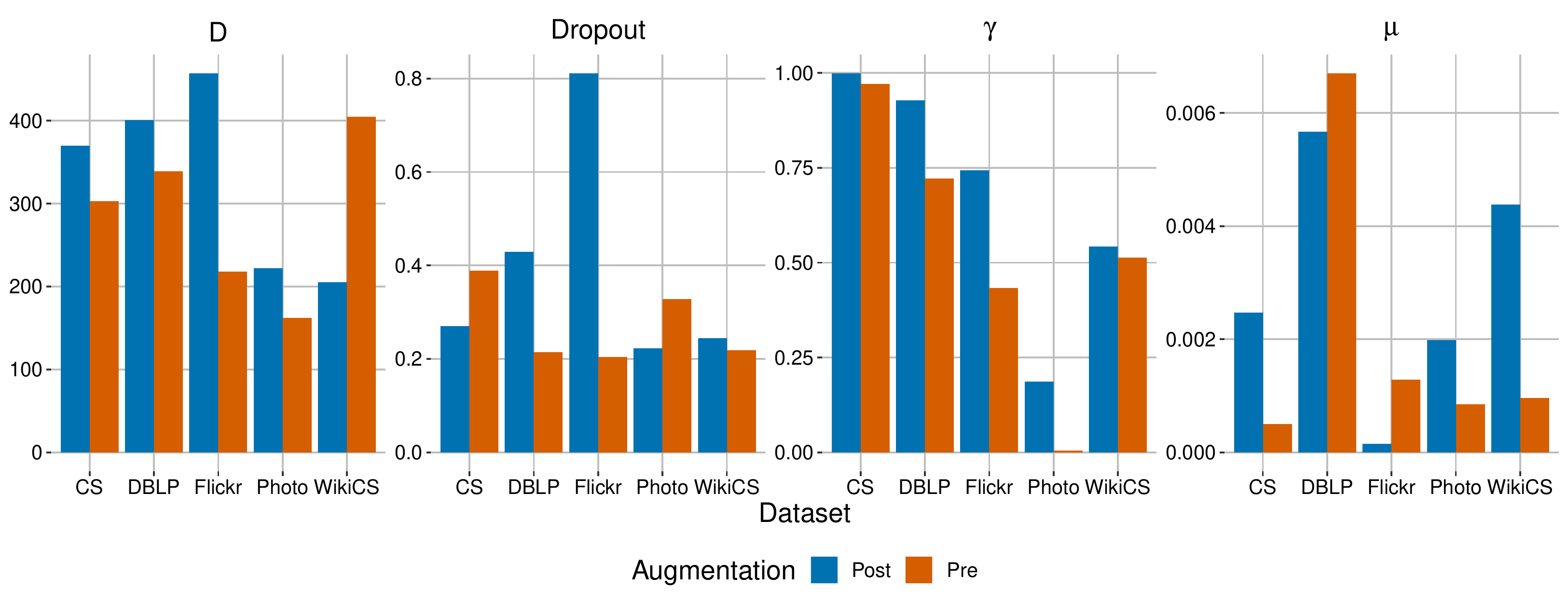}
    \caption{Tuned values of the hyperparameters of \model~ for the pre and post augmentation techniques. $D$ is the number of features after augmentation and the GNN encoder for pre and post augmentations, respectively. $\gamma$ the weight of the constraint in Eq.~\ref{eq:final_objective}, and $\mu$ is the learning rate.}
    \label{fig:pre_vs_post_hyperparams}
\end{figure}

In terms of performance, both augmentation techniques give equivalent results.
The only difference is that the hyperparameters need to be tuned separately for each one.
In Fig.~\ref{fig:pre_vs_post_hyperparams} we show the final configuration of the hyperparameters that maximize prediction accuracy on the validation set, which are obtained using Bayesian optimization.
As can be seen from the figure, they converge to different values.

\begin{SCfigure}
    \centering
    \includegraphics[scale=0.5]{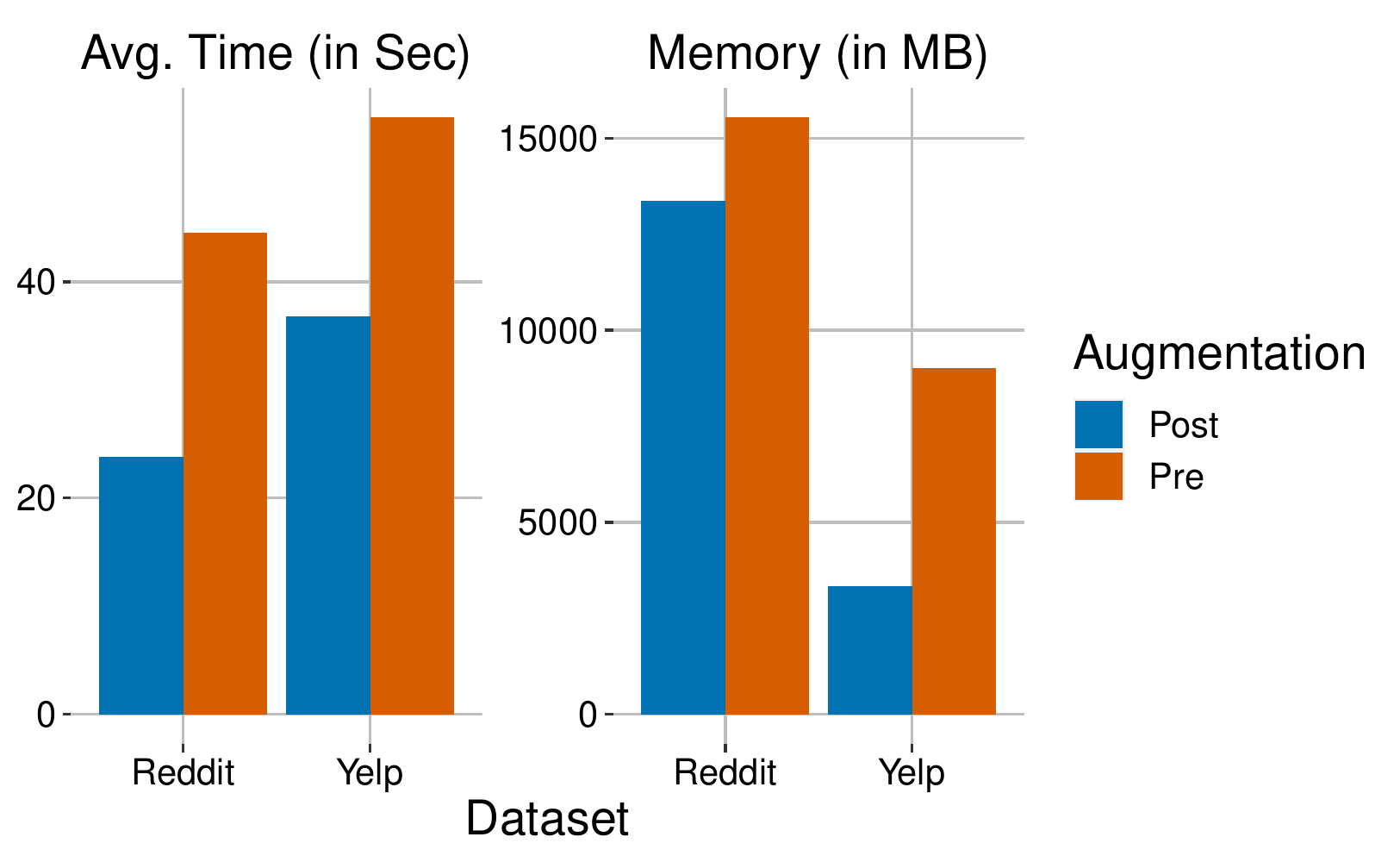}
    \caption{Analysis of Pre and Post Augmentation techniques in terms of memory usage and run time complexity (time to finish an epoch)}
    \label{fig:pre_vs_post_aug}
\end{SCfigure}

Since the main motivation for introducing the post-augmentation is efficiency, in Fig.~\ref{fig:pre_vs_post_aug} we show the memory usage and run time (to finish an epoch) required by these variants.
As anticipated, the post-augmentation is significantly faster and has a lot less memory overhead.


\begin{figure}
    \centering
    \includegraphics[scale=0.57]{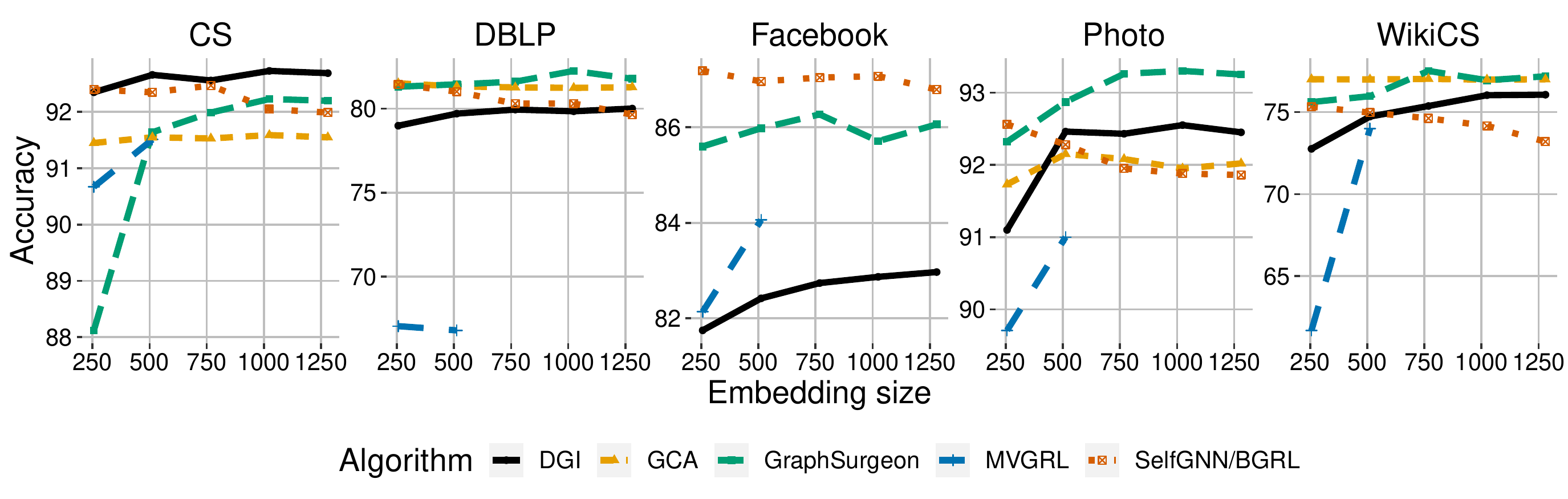}
    \caption{Effect of embedding size on~\model~and the~\gssl~baselines.}
    \label{fig:embedding_size_effect}
\end{figure}

\subsubsection{Symmetry vs. Asymmetry}
\begin{SCfigure}
    \centering
    \includegraphics[scale=0.5]{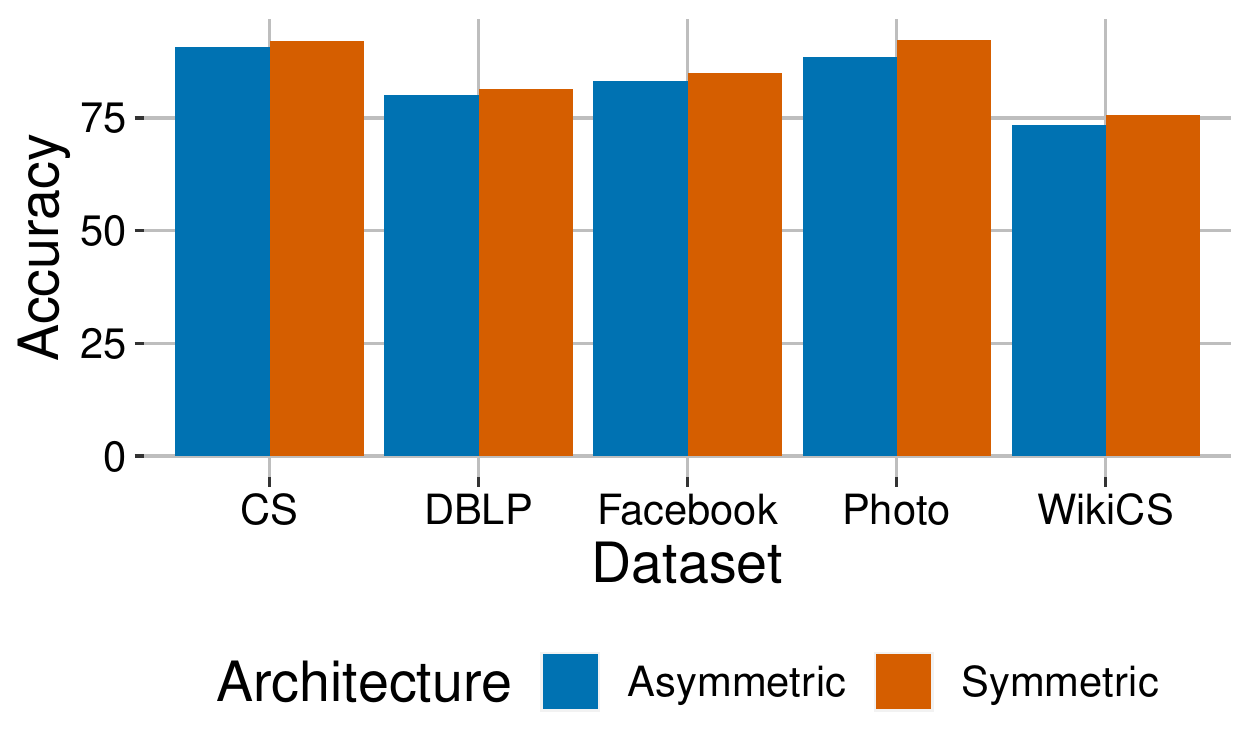}
    \caption{Comparison between \model's architecture~(Symmetric) and an Asymmetric architecture}
    \label{fig:sym_vs_asm}
\end{SCfigure}

\model's~ architecture can easily be replaced by an asymmetric one. 
For this reason, we create asymmetry just by adding a prediction head on the left-network and inserting batch norm in the GNN encoder~\cite{DBLP:journals/corr/abs-2103-14958,DBLP:journals/corr/abs-2103-03230,thakoor2021bootstrapped}.
We update the parameters of both the left and right networks using stochastic gradient descent.
As shown in Fig.~\ref{fig:sym_vs_asm}, the performance of the asymmetric architecture is slightly lower than the symmetric one.

\subsubsection{Convergence}
\begin{SCfigure}
    \centering
    \includegraphics[scale=0.6]{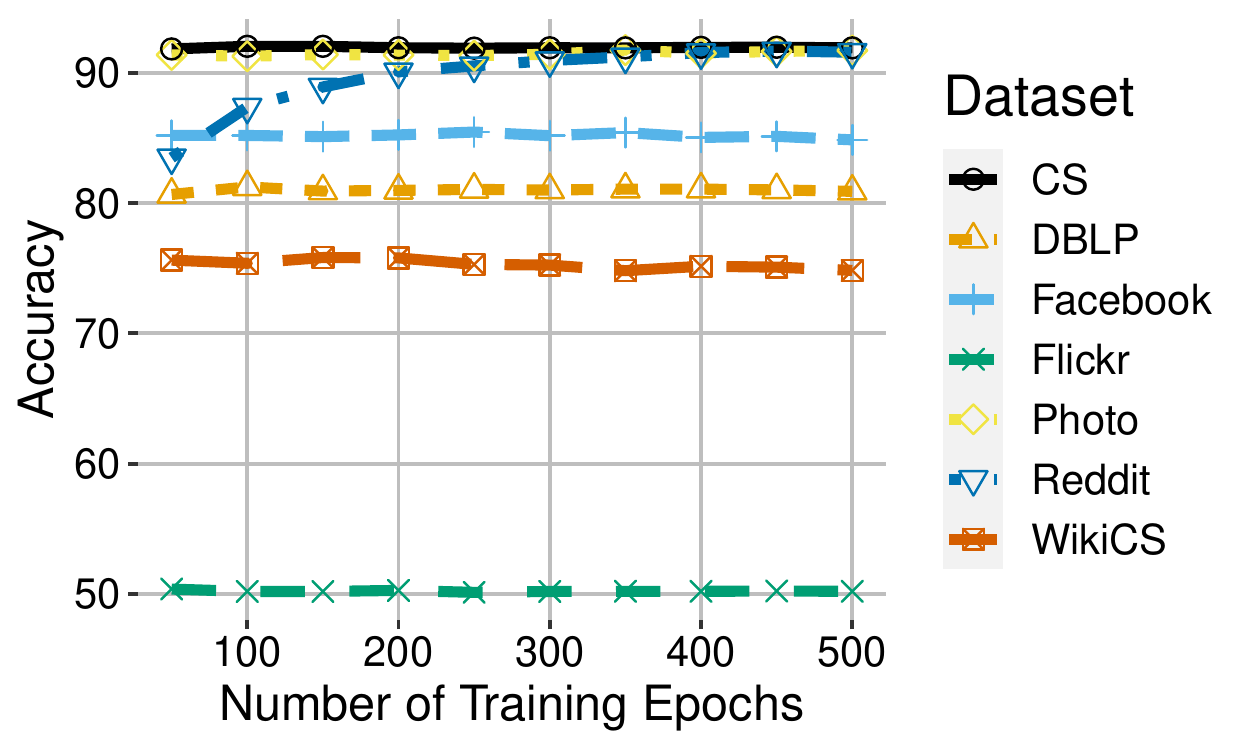}
    \caption{Analysis of the convergence of ~\model}
    \label{fig:convergence}
\end{SCfigure}

Recent studies~\cite{10.1145/3442381.3449802,DBLP:journals/corr/abs-2103-03230} have shown that \gssl~ methods usually require a large number of epochs (several thousand) to achieve a performance comparable to semi-supervised methods.
In Fig.~\ref{fig:convergence} we show ~\model's convergence, and usually, 50 epochs are sufficient to achieve comparable performance to semi-supervised methods.

\subsection{Implementation Details}
~\model~is implemented using PyTorch and PyTorch Geometric libraries.

For each augmentation head, we use a simple one-layer linear head.
To avoid overfitting, we use dropout in both heads.

For the GNN encoder we use two types of architectures, which are \gcn~\cite{kipf2017semisupervised} and \sage~\cite{hamilton2018inductive}.
A full-batch \gcn~is used for the small datasets, and a mini-batch GNN based on \sage~with neighborhood sampling~\cite{hamilton2018inductive} is used for the larger ones.
As stated earlier, one can substitute these with any other architecture as necessary.
A dropout is also added, and finally, we use a residual connection for the GNN encoder.

As indicated using the shape of the matrices in Listing~\ref{lst:surgeon_pseudo_code}, we use the output of the GNN encoder as the embedding of nodes for the pre-augmentation architecture and the output of the augmentation head for the post-augmentation case.

Although existing \gssl~methods~\cite{DBLP:journals/corr/abs-2103-03230,thakoor2021bootstrapped,thakoor2021bootstrapped,DBLP:journals/corr/abs-2103-14958} require different normalization strategies, such as Batch Norm and Layer Norm that is not necessary for~\model; as a result, no such normalization is used.

\section{Discussion and Conclusion}
\label{sec:conclusion}
In this paper, we propose a self-supervised graph representation learning method called~\model~based on the Siamese network.
Unlike prior methods that rely on heuristics for data augmentation, our method jointly learns the data augmentation with the representation (embedding) guided by the signal encoded in the graph.
By capitalizing on the flexibility of the learnable augmentations, we propose an alternative new strategy for augmentation, called post-augmentation, which happens after an encoding.
This is in contrast to the standard pre-augmentation strategy that happens before the encoding.
We also show that the alternative strategy significantly improves the scalability and efficiency of our model.

Furthermore, the method does not require explicit contrastive terms or negative sampling.
However, contrary to related studies with no contrastive terms, we employ a scalable principled constrained optimization inspired by Laplacian Eigenmaps, as opposed to engineering tricks to prevent trivial solutions.

We perform an extensive empirical evaluation of the proposed method using 14 publicly available datasets on three types of node classification tasks.
Besides, we compare the method with strong SOTA baselines, six semi-supervised GNNs, and five self-supervised GNNs.
Our finding shows that~\model~ is comparable to the semi-supervised GNNs and on-par with the self-supervised ones.

In this study, we focus on learning augmentations for attribute signals.
However, one can also consider topological augmentations and we shall explore this in future work.
In addition, it is also important to learn to identify if a particular type (topological/attribute) or a combination thereof is relevant to a down stream task.
Finally, this study does not inspect and interpret the learned augmentations, and further study is necessary to shed more insight regarding what is learned.







\bibliographystyle{ACM-Reference-Format}
\bibliography{main}










\end{document}